\documentclass[journal,twoside,web]{ieeecolor}
\usepackage{lcsys}
\usepackage{cite}
\usepackage{bm}
\usepackage{amsmath,amssymb,amsfonts}
\usepackage{algorithmic}
\usepackage{graphicx}
\usepackage{mathtools}
\usepackage{textcomp}
\usepackage{hyperref}
\def\BibTeX{{\rm B\kern-.05em{\sc i\kern-.025em b}\kern-.08em
    T\kern-.1667em\lower.7ex\hbox{E}\kern-.125emX}}
\begin{document}
\title{State-Space Kolmogorov Arnold Networks for Interpretable Nonlinear System Identification}
\author{Gonçalo G. Cruz, Balázs Renczes, Mark C. Runacres and Jan Decuyper
\thanks{\textbf{This article has been accepted for publication in IEEE Control Systems Letters. Citation information: DOI 10.1109/LCSYS.2025.3578019. For the publisher’s version and full citation
details see: \url{https://doi.org/10.1109/LCSYS.2025.3578019}}}
\thanks{Manuscript received March 17, 2025; revised April 21, 2025; accepted
June 2, 2025. Date of publication June 09, 2025.}
\thanks{This work was funded by the Strategic Research Program SRP60 of the Vrije Universiteit Brussel. }
\thanks{G.G. Cruz, M. C. Runacres and J. Decuyper are with the Faculty of Engineering
Technology, Vrije Universiteit Brussel, 1050 Brussel, Belgium
        (e-mail: goncalo.granjal.cruz@vub.be;mark.runacres@vub.be; jan.decuyper@vub.be)}
\thanks{B. Renczes is with the 
Department of Artificial Intelligence and Systems Engineering, Budapest University of Technology and Economics, Budapest, Hungary
        (e-mail: renczes@mit.bme.hu)}
\thanks{{\textcopyright 2025 IEEE. All rights reserved, including rights for text and data mining and training of artificial intelligence and similar technologies. Personal use is permitted,
but republication/redistribution requires IEEE permission.
}}
}

\pagestyle{empty} 
\maketitle
\thispagestyle{empty}

\begin{abstract}
While accurate, black-box system identification models lack interpretability of the underlying system dynamics. 
This paper proposes State-Space Kolmogorov-Arnold Networks (SS-KAN) to address this challenge by integrating Kolmogorov-Arnold Networks within a state-space framework. 
The proposed model is validated on two benchmark systems: the Silverbox and the Wiener-Hammerstein benchmarks.
Results show that SS-KAN provides enhanced interpretability due to sparsity-promoting regularization and the direct visualization of its learned univariate functions, which reveal system nonlinearities at the cost of accuracy when compared to state-of-the-art black-box models, highlighting SS-KAN as a promising approach for interpretable nonlinear system identification,
balancing accuracy and interpretability of nonlinear system dynamics.
\end{abstract}


\section{Introduction}
\label{sec:introduction}

\IEEEPARstart{S}{ystem} identification, the process of building mathematical models from observed data, is a fundamental discipline in engineering and control. Accurate system models are useful for tasks ranging from controller design and performance optimization to fault detection and system analysis. Linear system identification techniques have provided a robust and well-understood framework for modeling linear systems \cite{Pintelon2012SystemApproach}. However, real-world engineering systems are often nonlinear, and relying solely on linear models can lead to inadequate performance or even instability, particularly when operating in extended regimes or encountering complex dynamics. 
While black-box nonlinear system identification methods, especially those leveraging deep neural networks, have demonstrated impressive accuracy in capturing complex nonlinear behaviors \cite{Beintema2021}, as highlighted in comprehensive reviews of recent advancements and challenges in the field \cite{Pillonetto2025DeepSurvey}, a critical limitation remains: their inherent lack of interpretability. Black-box models lack transparency between model structure and system physics by representing dynamics in latent spaces with a high number of parameters, thus limiting the understanding of the identified system's behavior. 
Existing techniques like polynomial decoupling have been proposed to address this challenge by simplifying complex nonlinearities into more understandable forms, aiming to extract structured models from black-box representations \cite{Decuyper2019DecouplingModels,Decuyper2021RetrievingDecoupling}, though these might be constrained by the initial black-box model's structure. Other authors have explored augmenting physical models \cite{Liu2024Physics-GuidedNetworks} and imposing prior physical information \cite{Moradi2023Physics-InformedModels}, reinforcing  model transparency and physical consistency.

To address the interpretability challenge, Kolmogorov-Arnold Networks (KANs) \cite{liu2025kan} have emerged as an alternative, expressing any multivariate continuous function as a sum of compositions of learnable univariate functions. This structured decomposition makes KANs more interpretable than traditional black-box models, being explored in various scientific machine learning applications, potentially enabling hidden physics discovery \cite{Koenig2024,pal2024kanmultkanphysicsinformedsplinefitting}, though practical challenges are still under investigation \cite{Shukla2024ANetworks}.

In the context of system identification, KANs offer multiple advantages for modeling nonlinear systems, that are actively being studied \cite{cherifi2025nonlinearporthamiltonianidentificationinputstateoutput}.
Their structured representation, based on univariate functions, aims to capture complex nonlinear dynamics while simultaneously enhancing interpretability, making them attractive for modeling unknown system behaviors from input-output data.
Moreover, the functional decomposition inherent in KANs offers the potential to extract meaningful insights from identified models, such as identifying dominant input variables influencing system behavior. This transparency can be valuable for understanding the identified system and potentially for ensuring physical consistency in the learned model.

In this paper, we propose State-Space Kolmogorov Arnold Networks (SS-KAN), an approach
that integrates KANs into a state-space structure. Our primary contributions are the development of this architecture, the demonstration of its trade-off between accuracy and enhanced interpretability through the visualization of learned univariate functions, and its successful application to complex benchmark systems.

\section{Kolmogorov-Arnold Networks}

KANs are an alternative neural network architecture inspired by the Kolmogorov-Arnold representation theorem, 
which states that any continuous multivariate function $f(\mathbf{x}):[0,1]^n \mapsto \ \mathbb{R}$ can be expressed as a sum of compositions of univariate functions:
\begin{equation}
f(\mathbf{x})=\sum_{q=1}^{2 n+1} \Phi_q\left(\sum_{p=1}^n \phi_{q, p}\left(x_p\right)\right)
\end{equation}
where $\phi_{q,p}:[0, 1]\mapsto \mathbb{R}$ are ``inner'' univariate functions applied to individual input variables $x_p$, and $\Phi_{q}: \mathbb{R}\mapsto \mathbb{R}$ are ``outer'' univariate functions applied to the sum of these inner function outputs.
This essentially means that complex functions can be decomposed into sums of simpler, one-dimensional transformations.

Extending this concept, a KAN \cite{liu2025kan} structures these univariate functions sequentially into $L$ layers:
\begin{equation}
\operatorname{KAN}(\mathbf{x})=\left(\Phi_{L-1} \circ \Phi_{L-2} \circ \cdots \circ \Phi_1 \circ \Phi_0\right)(\mathbf{x})
\end{equation}
where $\circ$ denotes function composition and each layer function $\Phi_l$ consists of learnable univariate functions that connect to the next layer $l+1$.

Considering layer $l$ has $n_l$ nodes corresponding to outputs from the previous layer, and layer $l+1$ has $n_{l+1}$ nodes, then $\Phi_l$ can be expressed as a matrix of learnable univariate functions, $\phi_{l, i, j}(\cdot)$:
\begin{equation}
\Phi_l=\left(\begin{array}{cccc}
\phi_{l, 1,1}(\cdot) & \phi_{l, 1,2}(\cdot) &\!\! \cdots & \! \phi_{l, 1, n_l}(\cdot) \\
\phi_{l, 2,1}(\cdot) & \phi_{l, 2,2}(\cdot) &\! \!\cdots & \!\phi_{l, 2, n_l}(\cdot) \\[-1pt]
\vdots & \vdots & & \vdots \\[-1.5pt]
\phi_{l, n_{l+1}, 1}(\cdot) & \phi_{l, n_{l+1}, 2}(\cdot) &\!\! \cdots &\! \phi_{l, n_{l+1}, n_l}(\cdot)
\end{array}\right)
\end{equation}
where each univariate function $\phi$ is parameterized as a sum of a SiLU residual activation function and a linear combination of
B-spline functions, $\phi(x) = w_b SiLU(x) + w_s \sum_i c_i B_i(x)$ where the coefficients $c_i$ of the B-splines, as well as the scaling factors $w_b$ and $w_s$, are learnable parameters.

Unlike traditional neural networks with fixed nodal activations and learnable scalar weights on edges, KANs feature learnable univariate activation functions on the edges and summations at the nodes.
This can reduce the number of required parameters while simultaneously increasing interpretability, as the learned shapes of these edge functions can be visualized, making KANs an attractive option for interpretable nonlinear system identification.

\section{State-Space Kolmogorov-Arnold Networks}

Building upon the motivation for interpretable system identification and drawing inspiration from the structural advantages of state-space models and the interpretability of KANs, we propose State-Space Kolmogorov-Arnold Networks (SS-KAN).  This approach aims to combine the strengths of both methodologies: the physically meaningful structure of state-space representations with the function approximation and interpretability capabilities of KANs.

A general nonlinear dynamical system in a discrete-time state-space form is written as:
\begin{equation}
\begin{aligned}
\label{eq:ss-general}
 & \mathbf{x}(k+{1})=\mathbf{A x}(k)+\mathbf{B u}(k)+\mathbf{f}(\mathbf{x}(k), \mathbf{u}(k)) \\
& \mathbf{y}(k)=\mathbf{C x}(k)+\mathbf{D u}(k)+\mathbf{g}(\mathbf{x}(k), \mathbf{u}(k))
\end{aligned}
\end{equation}
where $\mathbf{x}(k) \in \mathbb{R}^{n_x}$ represents the $n_x$-dimensional state vector at time step $k$, $\mathbf{u}(k) \in \mathbb{R}^{n_u}$ is the $n_u$-dimensional forcing input, and $\mathbf{y}(k) \in \mathbb{R}^{n_y}$ denotes the $n_y$-dimensional observable output.  The matrices $\mathbf{A} \in \mathbb{R}^{n_x \times n_x}$ and $\mathbf{B} \in \mathbb{R}^{n_x \times n_u}$ are the discrete-time state and input matrices respectively, representing the linear part of the state transition. The function $\mathbf{f}: \mathbb{R}^{n_x} \times \mathbb{R}^{n_u} \rightarrow \mathbb{R}^{n_x}$ is a nonlinear function describing nonlinear state transitions. Similarly, $\mathbf{C} \in \mathbb{R}^{n_y \times n_x}$ and $\mathbf{D} \in \mathbb{R}^{n_y \times n_u}$ are the output and direct feedthrough matrices, and $\mathbf{g}: \mathbb{R}^{n_x} \times \mathbb{R}^{n_u} \rightarrow \mathbb{R}^{n_y}$ is a nonlinear function for the output mapping.
While traditional black-box system identification might directly model the entire nonlinear system using a large neural network, SS-KAN aims to retain the linear state-space structure and enhance interpretability by modeling the nonlinear functions $\mathbf{f}(\cdot)$ and $\mathbf{g}(\cdot)$. This explicit separation simplifies the complexity of the functions KANs must approximate, improving training stability and allowing for stable linear initialization, which is a common approach in system identification and naturally reduces initialization sensitivity inherent in KANs.

\subsection{SS-KAN model structure}

To formalize the SS-KAN architecture, we propose to replace the unknown nonlinear functions $\mathbf{f}(\cdot)$ and $\mathbf{g}(\cdot)$ in (\ref{eq:ss-general}) with KANs. This leads to the SS-KAN model equations:
\begin{equation}
\begin{aligned}
&\mathbf{x}(k+{1})=\mathbf{A x}(k)+\mathbf{B u}(k)+\text{KAN}_f(\mathbf{x}(k),\mathbf{u}(k)) \\
&\mathbf{y}(k)=\mathbf{C x}(k) + \mathbf{D u}(k)+\text{KAN}_g(\mathbf{x}(k),\mathbf{u}(k))
\end{aligned} 
\label{eq:ss_kan}
\end{equation}
where $\text{KAN}_f: \mathbb{R}^{n_x} \times \mathbb{R}^{n_u} \rightarrow \mathbb{R}^{n_x}$ and $\text{KAN}_g: \mathbb{R}^{n_x} \times \mathbb{R}^{n_u} \rightarrow \mathbb{R}^{n_y}$.
In this work, we utilize the \textit{efficientkan} \cite{Blealtan2024Efficientkan} implementation of KANs.
The inputs to both $\text{KAN}_f$ and $\text{KAN}_g$ are the state vector $\mathbf{x}(k)$ and the input vector $\mathbf{u}(k)$.

The trainable parameters of the SS-KAN model, denoted by $\bm{\theta}$, are composed of the linear state-space matrices and the weights of the KAN:
\begin{equation}
\bm{\theta} = [\mathbf{A}, \mathbf{B}, \mathbf{C}, \mathbf{D}, \bm{\theta}_{\text{KAN}_f}, \bm{\theta}_{\text{KAN}_g}]{^T}\label{eq:params}
\end{equation}
where $\bm{\theta}_{\text{KAN}_f}$ and $\bm{\theta}_{\text{KAN}_g}$ represent the sets of trainable parameters within the KANs used for the state transition and output mapping nonlinearities, respectively.

\subsection{Cost Function and Optimization}

The SS-KAN model parameters (\ref{eq:params}) are trained by minimizing a cost function that balances model accuracy with interpretability: 
\begin{equation}
\begin{aligned}
\mathcal{L}(\bm{\theta})  &=  \frac{1}{N} \left\lVert \mathbf{y}_{SS-KAN} - \mathbf{y}_{data} \right\rVert_2^2 \\ &+
\lambda_{L2} \left( \left\lVert \mathbf{A} \right\rVert_F^2 + \left\lVert \mathbf{B} \right\rVert_F^2 + \left\lVert \mathbf{C} \right\rVert_F^2 + \left\lVert \mathbf{D} \right\rVert_F^2 \right) \\ &
+ \lambda_{L1}\left(\lVert\bm{\theta}_{\text{KAN}_f}\rVert_1+\lVert\bm{\theta}_{\text{KAN}_g}\rVert_1\right)
\label{eq:loss}
\end{aligned}
\end{equation}
Here, the first term represents the mean squared $L2$ norm error, quantifying the discrepancy between the predicted output $\mathbf{y}_{SS-KAN}$ and the available $\mathbf{y}_{data}$ over $N$ data points. The second term is an $L2$ regularization penalty with the Frobenius norm applied to the linear state-space matrices ($\mathbf{A}, \mathbf{B}, \mathbf{C}, \mathbf{D}$), with $\lambda_{L2}$ controlling its strength to improve generalization of the linear components and prevent overfitting. The third term is an $L1$ regularization penalty on the KAN parameters ($\bm{\theta}_{\text{KAN}_f}$ and $\bm{\theta}_{\text{KAN}_g}$), with $\lambda_{L1}$ controlling its strength to promote sparsity in the activation functions thus increasing the model interpretability.
In this work, for the internal KAN architecture, we utilize cubic B-splines defined on a 5-point grid 
for each univariate function and use two layers to showcase benchmark interpretability. Increasing grid points can enhance detail but adds parameters, while deeper KANs might be needed for highly complex systems.

To optimize the model parameters $\bm{\theta}$, we utilize the AdamW optimizer \cite{Loshchilov2017DecoupledRegularization} with conservative initial learning rates, as this was found to stabilize convergence.
The training process employs batch optimization which was found to speedup convergence. The training dataset is divided into empirically chosen $B$ batches of size $N_{batch} = N/B$. The data is normalized before training to a [-1,1] range where the B-splines domain is initially defined, and is processed sequentially in time to maintain temporal dependencies within each epoch.

The performance of the SS-KAN model is quantitatively evaluated using the Root Mean Squared Error (RMSE). RMSE provides a measure of the average error magnitude between the predicted output $\mathbf{y}_{SS-KAN}$ and the measured data $\mathbf{y}_{data}$.  For a dataset with $N$ data points, RMSE is defined as:
\begin{equation}
\text{RMSE} = \sqrt{\frac{1}{N} \sum_{k=1}^{N} \left(\mathbf{y}_{SS-KAN}(k) - \mathbf{y}_{data}(k)\right) ^2}
\label{eq:rmse_definition}
\end{equation}

\subsection{On the connection with decoupled state-space models}  
A key aspect of regaining interpretability in KANs lies in their reliance on an additive structure of univariate functionals. Since univariate functions can be easily visualized, this often leads to valuable insights into nonlinear relationships. Building on similar reasoning, so-called decoupled functions have also been proposed. In this approach, nonlinearity is constrained within a set of so-called univariate ``branches'', effectively decoupling the multivariate relationship. Decoupled functions are defined as $f(\mathbf{x})= \mathbf{W} \mathbf{g} \left(\mathbf{V}^T \mathbf{x} \right)$,
with univariate functions $g_i(z_i): \mathbb{R} \mapsto \mathbb{R}$ of linear forms $z_i \coloneqq \mathbf{v}^{\top}_i \mathbf{x}$, and $i=1,\ldots, r$. Here, $r$ is the number of branches, and $\mathbf{W}$ and $\mathbf{V}$ are linear transformation matrices. It has been demonstrated \cite{Decuyper2021DecouplingApproach} that a broad class of multivariate functions can be accurately approximated in a decoupled form. 
A notable distinction is that KANs use regularization to promote sparsity and model simplicity, while in decoupled functions, model complexity is usually controlled through the parameter $r$. Beyond visualizing the nonlinearities, further insight may be gained by examining the inputs to the nonlinear components, particularly in systems where a dominant nonlinear term involves a specific physical state variable (e.g., a hardening spring in mechanical systems, see section \ref{s:silverbox}). In this regard \cite{Decuyper2021RetrievingDecoupling} demonstrated that single-branch decoupled functions, when embedded in state-space models, could yield physically meaningful intermediate variables $z_i$. However, this physical interpretability is not preserved in the state variables themselves, due to the mixing introduced by the linear transformation matrix $\mathbf{V}$. KANs, on the other hand, guided towards a lean network through regularization, may naturally revert to a structure where only one of the input variables drives the nonlinearity. Since SS-KANs take the state variables as input, and under the assumption of a well-approximated nonlinearity, this structure may help preserve or even promote physical interpretability of the state variables themselves, particularly for systems governed by a single dominant nonlinear term.

\section{Silverbox test case - Duffing oscillator}
\label{s:silverbox}

\begin{figure*}[htbp]
    \centering    \includegraphics[width=\textwidth,trim={0.0cm 1.2cm 0.0cm 0.5cm}]{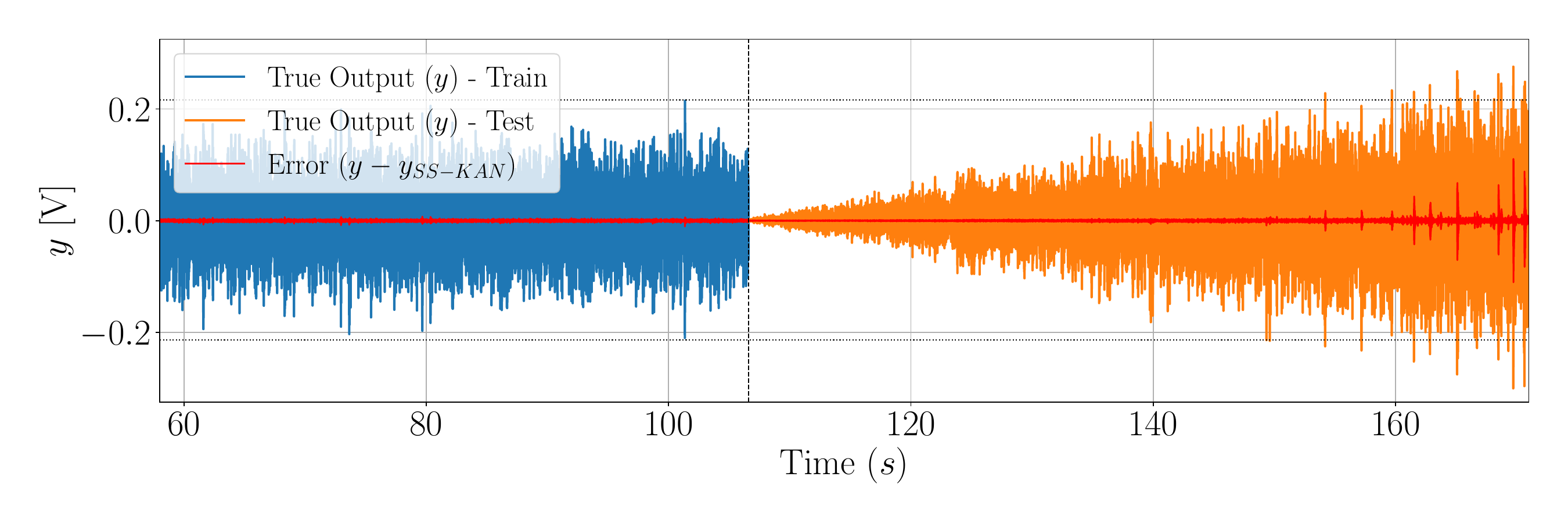}
    \caption{Time-Domain Performance of SS-KAN Model on Silverbox Benchmark. The blue and orange lines show the true output signal for the (partial) training and test sets, respectively. The red line represents the simulation error ($y - y_{SS-KAN}$) across both datasets. The vertical dashed line indicates the split between the training and testing data. The horizontal dotted lines indicate the amplitude extrapolation bound where the error increases.}
    \label{fig:silverbox_error}
\end{figure*}

To evaluate the effectiveness of the SS-KAN model, we consider the well-known Silverbox benchmark system \cite{Marconato2012IdentificationModels}, which can be viewed as an electronic version of the forced Duffing oscillator, a second-order nonlinear system with a cubic nonlinearity. The Duffing oscillator is a widely recognized benchmark in nonlinear system identification due to its well-characterized nonlinear behavior and relevance to many physical systems, particularly in mechanics and structural dynamics.  The true equation of motion for the Duffing oscillator is given by:
\begin{equation}
    m \ddot{x} + c \dot{x} + k x + \alpha x^3 = u(t)
    \label{eq:duffing_equation}
\end{equation}
where, $x$ represents the displacement, $\dot{x}$ and $\ddot{x}$ are its time derivatives, $m$ is the mass, $c$ is the damping coefficient, $k$ is the linear stiffness, $\alpha$ is the cubic nonlinear stiffness, and $u(t)$ is the external forcing input. The cubic term $\alpha x^3$ is the key nonlinear element that we aim to identify and interpret using our SS-KAN approach in a pure system identification approach where the true parameters of the Duffing oscillator (m, c, k, $\alpha$) are unknown.

The data for the Silverbox benchmark is obtained from the Nonlinear Benchmark repository \cite{shoukens_github_nonlinear}. It consists of a training set of $N_{train}=65000$ samples and a test set of $N_{test}=40000$ samples. 
The training data comprises a collection of random phase multisines with a constant amplitude in a frequency range up to 200 Hz. The test data consists of filtered Gaussian noise with a linearly increasing amplitude, including extrapolation regions beyond the training range. 

\subsection{SS-KAN Model Setup}
For state-space representation, we define the state vector $\mathbf{x} = \begin{bmatrix} x & \dot{x} \end{bmatrix}^T$, where $x$ represents position and $\dot{x}$ velocity. 

The discrete-time SS-KAN model for the Silverbox benchmark utilizes a state vector of dimension $n_x = 2$ and a scalar input $n_u = 1$ and output $n_y = 1$, making it a single input single output (SISO) test case.
The KAN for the state transition, $\text{KAN}_f$, consists of 2 layers with the hidden layer size $n_l =2$, while the KAN for the output mapping, $\text{KAN}_g$, is set to zero ($\text{KAN}_g = 0$), implying a purely linear output mapping in this SS-KAN configuration. 
The data are scaled to the range $[-1, 1]$ to align with the grid domain of the B-spline basis functions used in the KAN implementation. 
The linear state-space matrices ($\mathbf{A}, \mathbf{B}, \mathbf{C}, \mathbf{D}$) are initialized to represent a stable, weakly damped linear system close to an identity mapping. The regularization penalties are applied with $\lambda_{L1} = \lambda_{L2} = 10^{-4}$.
The AdamW optimizer is run with a learning rate set to $10^{-3}$ and a batch size $N_{batch} = 64$ for 100 epochs ($\approx 1$ hour) in a M4 Pro (10 CPU cores at 4.5GHz).

\subsection{Results}

The RMSE on the test data obtained on the Silverbox benchmark with the proposed SS-KAN model is compared against several established baseline models reported in the literature and summarized in Table \ref{tab:rmse_comparison}.
\begin{table}[h]
\centering
\caption{RMSE Comparison on Silverbox Test Data}
\label{tab:rmse_comparison}
\addtolength{\tabcolsep}{-0.01cm}
\begin{tabular}{llll}
\hline
\textbf{Model} & \textbf{Model Type} & \textbf{RMSE} [V] & \textbf{Train Time}\\
\hline
BLA \cite{Marconato2012IdentificationModels} & Linear Black-Box & 0.0135 & $\sim$ seconds \\
PNLSS \cite{Paduart2010} & Nonlinear Black-Box & 0.0003 & $\sim$ hours\\
Deep Encoder \cite{Beintema2021} & Nonlinear Black-Box & 0.0014 & $\sim$ days\\
SS-KAN & Nonlinear Grey-Box & 0.0039 & $\sim$ hours\\
\hline
\end{tabular}
\end{table}

The proposed SS-KAN model achieves a Test RMSE of $\mbox{0.0039 V}$ on the Silverbox benchmark.  While the mature polynomial nonlinear state-space (PNLSS) model demonstrates the lowest RMSE, achieving the highest quantitative accuracy on this benchmark, linked to the underlying system's polynomial nature, SS-KAN still exhibits a significantly lower RMSE – approximately one order of magnitude – compared to the Best Linear Approximation (BLA). Compared to the Deep Enconder approach, SS-KAN's RMSE is slightly higher.

However, our SS-KAN implementation, not only preserves the original state dimensions but has fewer degrees of freedom due to the simpler chosen architecture, training effectively in 100 epochs, requiring only around one hour of computation time on a modern laptop. 
This contrasts with the reported multi-day training for the Deep Encoder approach but is comparable to PNLSS models which achieve high accuracy but have training times that are dependent on specific model complexity (e.g., polynomial degree used).
While not achieving the absolute lowest RMSE, the quantitative performance of SS-KAN in Table \ref{tab:rmse_comparison} demonstrates a strong balance between accuracy and efficiency. 
A qualitative analysis of the time-domain response offers further insights regarding its ability to track the system's dynamic behavior over time.  
Fig. \ref{fig:silverbox_error} presents the time-domain absolute error between the SS-KAN model predicted output and the measured output for part of the training and testing datasets.

The simulation error (red line) remains  small throughout the majority of the time series, indicating accurate tracking of the system's dynamics by the SS-KAN model. 
A slight increase in simulation error is observed within the extrapolation region (horizontal dotted lines) in the test data. Despite this, the overall error magnitude remains low, demonstrating good generalization and suggesting that the nonlinear dynamics have been effectively identified by the KAN.
The low simulation error visualized in Fig. \ref{fig:silverbox_error} supports the quantitative RMSE results presented in Table \ref{tab:rmse_comparison}, confirming the SS-KAN model's ability to accurately capture the nonlinear dynamics of the Silverbox benchmark in the time domain. 

After showcasing the SS-KAN's ability to capture the system's dynamics over time, we now focus on the interpretability advantage of the model.
Since $\text{KAN}_f$ in the SS-KAN model (\ref{eq:ss_kan}) learns the nonlinear state updates, visualizing its output as a function of the state and input can reveal how the model represents the system's nonlinear dynamics, especially since the KAN preserves the function inputs without mixing, unlike the decoupling approach discussed above.

\begin{figure}[!htbp]
\centering
\hspace*{-0.7cm}                                                           
    \includegraphics[width=0.8\columnwidth,trim={0.0cm 0.9cm 0.0cm 0.5cm}]{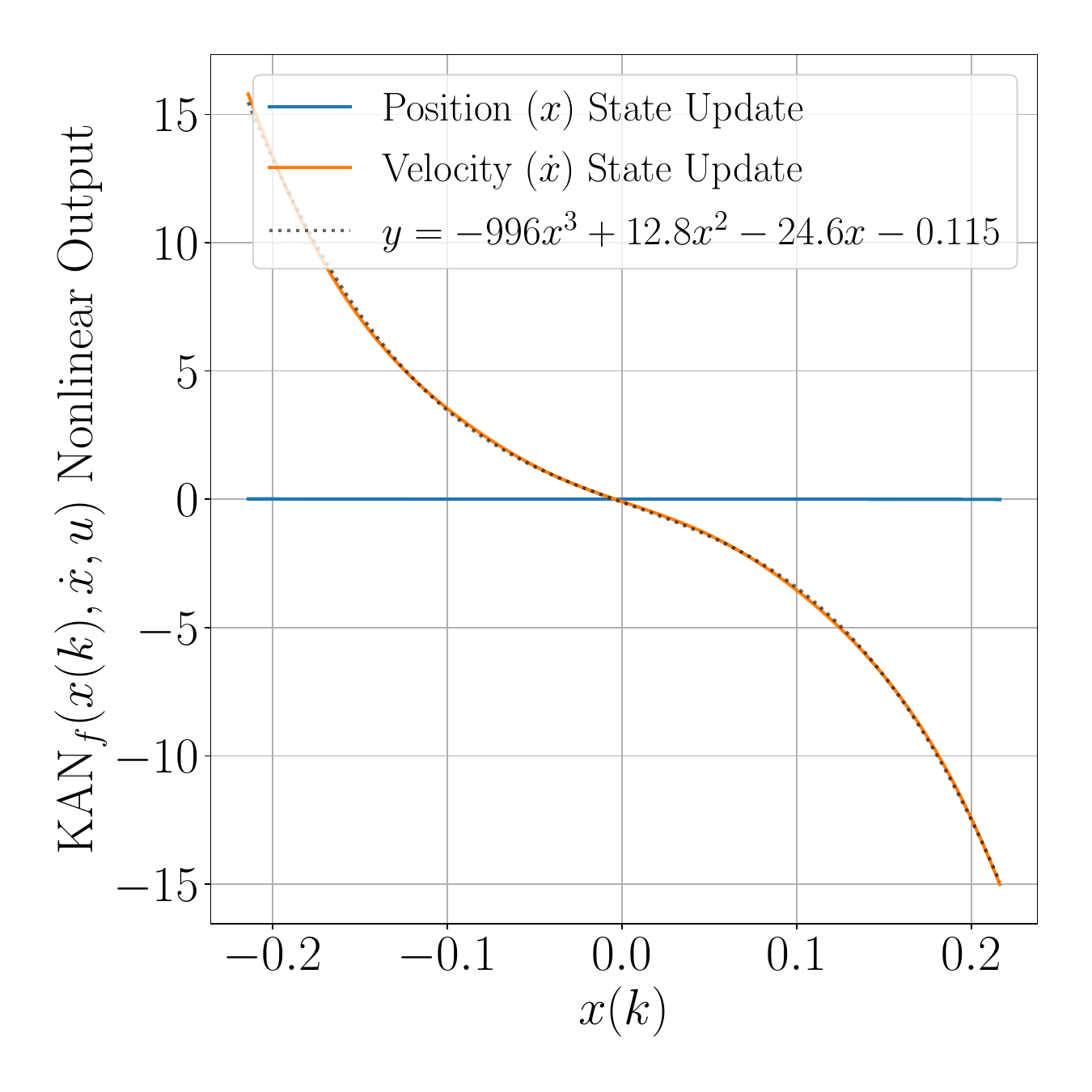}
    \caption{Learned {KAN}$_f({x}(k),\dot{x} , u)$ nonlinear functions by varying the position state variable ($x$) for both position $x$ (blue) and velocity $\dot{x}$ (orange) state updates. 
    The velocity $\dot{x}$ state update is well-approximated by $y \approx -996 x^3 + 12.8x^2 - 24.6x -0.115$ (dotted black). The dominant cubic term captures the Duffing oscillator's stiffness, while the smaller quadratic and linear terms reveal the KAN's ability to identify more subtle dynamic effects.}
    \label{fig:kan_output}
\end{figure}

For the Silverbox test case, we focus on visualizing the univariate functions within $\text{KAN}_f$ by fixing the velocity state ($\dot{x}$) and the input ($u$) to their mean values and then varying the position state ($x$) over the training data range.
As it can be seen in Fig. \ref{fig:kan_output}, this allows to isolate and visualize the direct influence of the displacement state variable on the learned nonlinear state updates.

The KAN function for the velocity $\dot{x}$ state update (orange) exhibits a cubic shape, indicating that SS-KAN effectively captures the cubic stiffness nonlinearity of the Duffing oscillator in the velocity state equation. A polynomial fit shows it to be well approximated by $y \approx -996 x^3 + 12.8x^2 - 24.6x -0.115$ (dotted black line). The dominant cubic term clearly indicates that SS-KAN effectively captures the cubic stiffness nonlinearity of the Duffing oscillator. The presence of a quadratic term suggests the model has identified secondary nonlinear effects, known to be potentially present in the Silverbox benchmark's physical realization. The residual linear dynamics are captured by the KAN, which is not constrained to model purely nonlinear effects and can thus account for linear components not fully represented by the global linear matrices. The constant term is minimal, suggesting a minor offset.
In contrast, the KAN function for the position state update (blue) remains near-zero, suggesting a negligible nonlinear contribution to the position state evolution. This analysis aligns with the expected nonlinear dynamics in (\ref{eq:duffing_equation}).
Moreover, the cubic shape of the KAN$_f$ velocity state update remains consistent even when the fixed values of the velocity state ($\dot{x}$) and input ($u$) for Fig. \ref{fig:kan_output} are varied across their respective ranges from minimum to maximum values observed in the training data. This highlights that the SS-KAN model correctly identifies the position state ($x$) as the dominant input variable driving the cubic nonlinearity, with the other inputs playing a negligible role in shaping this specific nonlinear behavior.
In parallel, a complementary analysis to further emphasize this point, omitted here, of fixing both the position state $x$ and input $u$, while varying the velocity state $\dot{x}$ and of fixing the both position and velocity states $\mathbf{x}$, while varying the input $u$, was performed. 
This analysis confirms the negligible magnitude of the univariate functions in both cases, reinforcing the interpretability strengths of the SS-KAN approach, revealing not only the functional shape of the dominant cubic nonlinearity but also the negligible influence of other input variables on the state updates. 

\section{Wiener-Hammerstein Test case}

\begin{figure}[htbp]
    \centering
    \includegraphics[width=0.8\columnwidth]{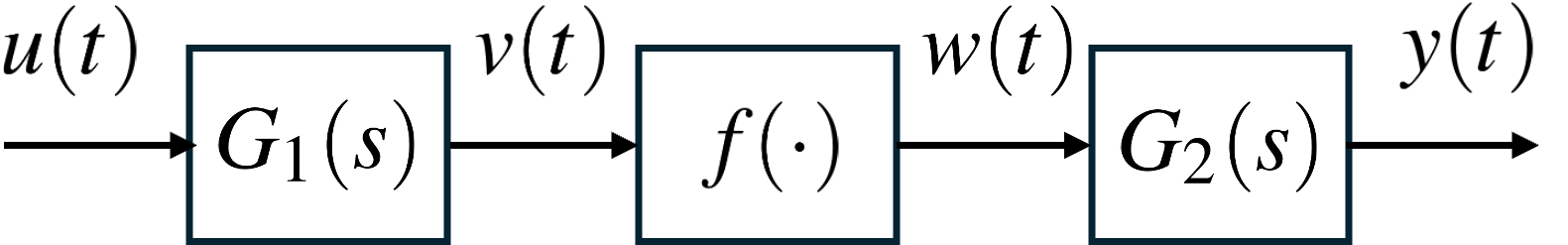}
    \caption{Schematic description of the Wiener-Hammerstein system.}
    \label{fig:wh_system}
\end{figure}

To further evaluate the generalizability of the proposed SS-KAN model, we consider the Wiener-Hammerstein benchmark system \cite{Schoukens2009Wiener-HammersteinBenchmark_full}, schematically represented in Fig. \ref{fig:wh_system}.
This benchmark presents a distinct identification challenge compared to the Silverbox. Instead of a localized nonlinearity within the state dynamics, the Wiener-Hammerstein system features a saturation-type nonlinearity (diode-resistor), $f(\cdot)$ between two third order linear dynamic blocks, $G_1(s)$ and $G_2(s)$.
This structure and the lack of direct access to internal states require the model to infer the system's nonlinear behavior only from the input-output signals, respectively, $u(t)$ and $y(t)$, making it a SISO system.
The dataset, generated with an electronic circuit excited by a filtered Gaussian excitation signal, consists of $N_{train}=80 000$ samples for training and $N_{test} =78 000$ samples for testing. 

\subsection{SS-KAN Model Setup}

To model the Wiener-Hammerstein structure within the SS-KAN framework, we adapt the general model equations (\ref{eq:ss_kan}) to explicitly represent the cascaded linear-nonlinear-linear blocks. The discrete-time state-space equations are given by:

\begin{equation}
\begin{aligned}
\mathbf{x}_1(k+1)&=\mathbf{A_1 x}_1(k)+\mathbf{B_1}u(k)\\
{v}(k)&=\mathbf{C_1 x}_1(k) + \mathbf{D_1}u(k)\\
 {w}(k) &= \text{KAN}({v}(k)) \\
\mathbf{x}_2(k+1)&=\mathbf{A_2 x}_2(k)+\mathbf{B_2}w(k)\\
{y}(k)&=\mathbf{C_2 x}_2(k) + \mathbf{D_2}w(k)
\end{aligned} 
\label{eq:ss_kan_WH}
\end{equation}
where, $\mathbf{x}_1(k) \in \mathbb{R}^{3}$ and $\mathbf{x}_2(k) \in \mathbb{R}^{3}$ represent the state vectors of the linear blocks.
The intermediate signal $v(k)$ represents the output of the $G_1(s)$ linear block and is the input to the static nonlinearity, modeled by $\text{KAN}: \mathbb{R} \rightarrow \mathbb{R}$, approximating the static diode-resistor nonlinearity. 
The output of the KAN then drives the $G_2(s)$ linear block.  The overall measured system output is denoted by $y(k)$.

The KAN architecture consists of 2 layers with a hidden layer size $n_l\!=\!15$. The dynamic grid update approach \cite{liu2025kan} adapts the spline grids based on the input activations range during training.
The linear matrices ($\mathbf{A_1}, \mathbf{B_1}, \mathbf{C_1}, \mathbf{D_1}, \mathbf{A_2}, \mathbf{B_2}, \mathbf{C_2}, \mathbf{D_2}$) are initialized using Chebyshev filter information from the benchmark reference.
The regularization penalties are $\lambda_{L1} \!=\! \lambda_{L2} \!= 10^{-4}$.
The AdamW optimizer is run with a decaying learning rate from $10^{-4}$ and a batch size $N_{batch} = 2048$ for 500 epochs ($\approx\! 6\!-\!8$ hours) using the same hardware as the Silverbox test case.

\subsection{Results}

Similar to the Silverbox test case, we quantitatively evaluate the performance of the SS-KAN model on the Wiener-Hammerstein benchmark by comparing the obtained test RMSE to literature, as summarized in Table \ref{tab:rmse_comparison_WH}.

\begin{table}[h]
\centering
\caption{RMSE Comparison on Wiener-Hammerstein Test Data}
\label{tab:rmse_comparison_WH}
\addtolength{\tabcolsep}{-0.01cm}
\begin{tabular}{llll}
\hline
\textbf{Model} & \textbf{Model Type} & \textbf{RMSE} [V] &\textbf{Train Time}\\
\hline
BLA \cite{Lauwers2009ModellingApproximation} & Linear Black-Box & 0.0562 & $\sim$ seconds\\
PNLSS \cite{Paduart2012IdentificationApproach} & Nonlinear Black-Box & 0.0004 & $\sim$ hours\\
Deep Encoder \cite{Beintema2021} & Nonlinear Black-Box & 0.0002 & $\sim$ days\\
SS-KAN & Nonlinear Grey-Box & 0.0114 & $\sim$ hours\\
\hline
\end{tabular}
\end{table}

\begin{figure}[htbp] 
\centering
\hspace*{-0.7cm}                                                  
\includegraphics[width=0.8\columnwidth,trim={0.0cm 0.9cm 0.0cm 0.9cm}]{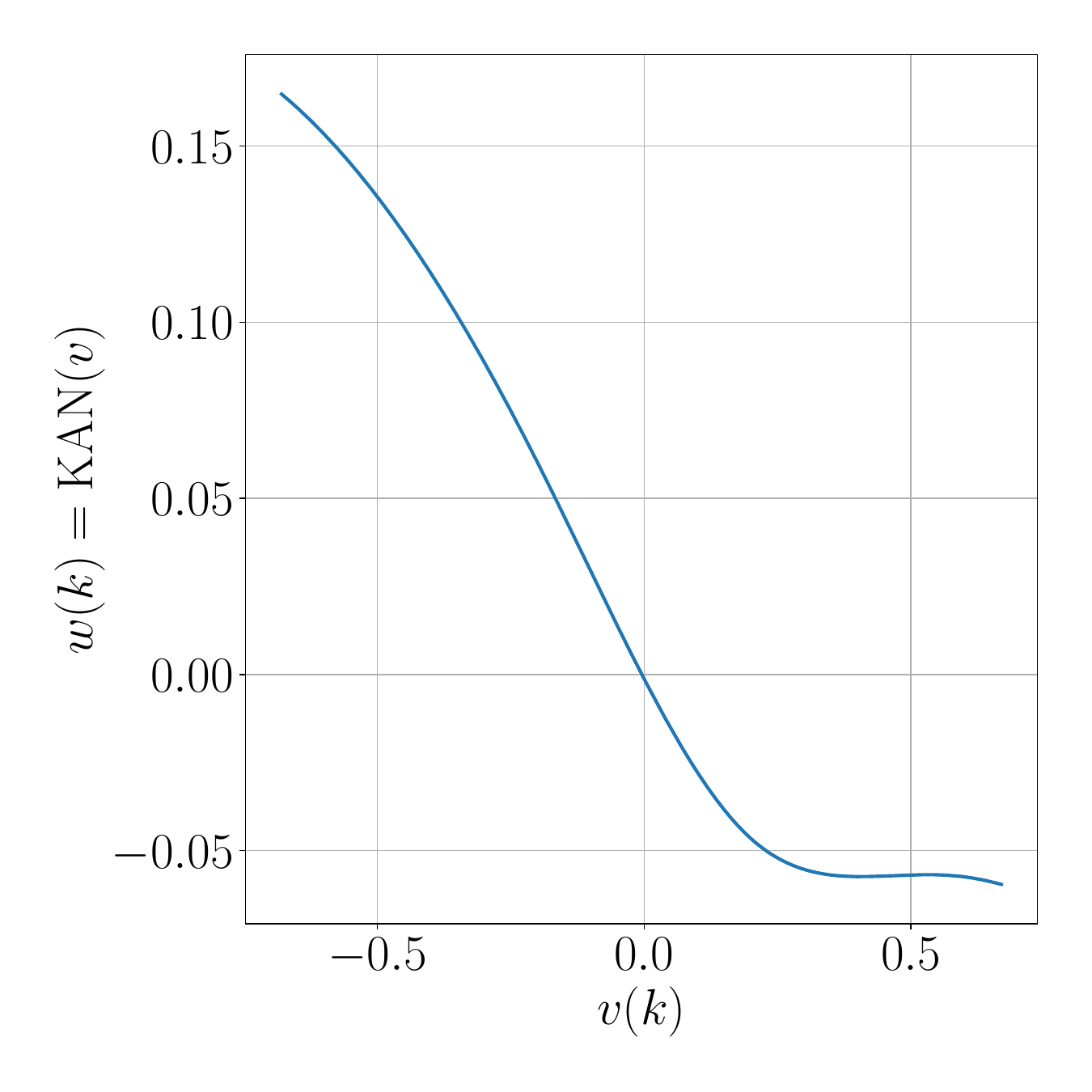}
    \caption{Learned KAN function for Wiener-Hammerstein nonlinearity. The x-axis represents the input to the KAN, the intermediate signal $v(k)$, and the y-axis represents the output of the KAN, $w(k)$. It exhibits a linear trend with saturation, directly reflecting the behavior of the diode-resistor nonlinearity.}
    \label{fig:kan_output_wh}
\end{figure}

The SS-KAN model achieves a Test RMSE of $\mbox{0.0114 V}$ on the Wiener-Hammerstein benchmark. 
As shown in Table \ref{tab:rmse_comparison_WH}, while SS-KAN outperforms the BLA, its RMSE is higher than that of the PNLSS and Deep Encoder models.
This highlights a trade-off between prioritizing SS-KAN interpretability against achieving the lowest possible quantitative error of black-box models.
In terms of computational efficiency, SS-KAN ($\sim$ 6-8 hours) is considerably faster than the Deep Encoder ($\sim$ days) and falls within a similar order of magnitude as PNLSS models ($\sim$ hours), whose specific training times are highly dependent on their complexity.
Fig.~\ref{fig:kan_output_wh} displays the $\text{KAN}$ learned nonlinear function of the Wiener-Hammerstein system (\ref{eq:ss_kan_WH}).
The KAN exhibits a linear trend with saturation. For lower input values, the function shows a near-linear trend, corresponding to the diode being "off" or non-conducting, resulting in a linear relationship (voltage divider behavior) in the electronic circuit. Beyond a certain input threshold, the function saturates and becomes nearly constant, representing the diode "turning on" and clamping the voltage to a saturation level, thus capturing the saturation nonlinearity of the diode-resistor circuit.
The physical interpretation of the learned nonlinear function, is a key strength of the SS-KAN approach, particularly in contrast to black-box models where such insights are often hidden in latent spaces or hyperparameters.

\section{Conclusion}
We introduced State-Space Kolmogorov-Arnold Networks (SS-KAN), a new approach for interpretable nonlinear system identification that integrates Kolmogorov-Arnold Networks into a state-space framework.
The analysis on the Silverbox and Wiener-Hammerstein benchmarks demonstrates that SS-KAN trades-off black-box quantitative accuracy for interpretability.
While SS-KAN does not reach the absolute lowest RMSE compared to highly flexible black-box models, the visualization of learned KAN functions within SS-KAN provides direct and physically meaningful insights into the system nonlinearities, revealing the cubic stiffness of the Duffing oscillator and the saturation characteristic of the Wiener-Hammerstein system.
This enhanced interpretability, achieved without sacrificing significant accuracy, is the key contribution of SS-KAN, offering a valuable tool for interpretable system identification.

\bibliographystyle{IEEEtran}
\bibliography{references_mend} 
\end{document}